\documentclass[letterpaper, 10 pt, conference]{ieeetran}
\def\IEEEtitletopspace{18pt}
\IEEEoverridecommandlockouts
\usepackage{times}

\usepackage[numbers]{natbib}
\usepackage{multicol}
\usepackage[bookmarks=true]{hyperref}
\usepackage{graphicx}

\usepackage{amsfonts}       %
\usepackage{nicefrac}       %
\usepackage{microtype}      %
\usepackage{amsmath}
\usepackage{amssymb}
\usepackage{gensymb}
\usepackage{multirow}
\usepackage{booktabs}
\usepackage{pbox}
\usepackage{array} 
\usepackage{hyperref}
\usepackage[ruled]{algorithm2e}
\usepackage{float}
\usepackage{longtable}

\renewcommand{\eqref}[1]{(\ref{#1})}

\newcommand{\subfig}[1]{\textit{#1}}

\newcommand{\ie}{\textit{i.e.}}
\newcommand{\eg}{\textit{e.g.}}
\newcommand{\etc}{\textit{etc.}}
\newcommand{\etal}{\textit{et~al.}}

\newcommand{\gdoom}{\mbox{G-DOOM}\xspace}

\makeatletter 
\newcommand\figcaption{\def\@captype{figure}\caption} 
\newcommand\tabcaption{\def\@captype{table}\caption} 
\makeatother

\def\BibTeX{{\rm B\kern-.05em{\sc i\kern-.025em b}\kern-.08em
    T\kern-.1667em\lower.7ex\hbox{E}\kern-.125emX}}
\begin{document}

\title{Learning Latent Graph Dynamics for Visual Manipulation of Deformable Objects}

\author{Xiao Ma, David Hsu,~\IEEEmembership{Fellow,~IEEE}, and Wee Sun Lee
\thanks{The authors are with the School of Computing, National University of Singapore. Email: \{\texttt{xiao-ma, dyhsu, leews\}@comp.nus.edu.sg}}
\thanks{This research is supported in part by the National Research Foundation (NRF), Singapore under its AI Singapore Program (AISG Award No: AISG2-RP-2020-016) and by the Agency of Science, Technology \& Research, Singapore, through the National Robotics Program (Grant No. 192 25 00054).}
}

\maketitle

\begin{abstract}
Manipulating  deformable objects, such as ropes and clothing, is a long-standing challenge in robotics, because of their large degrees of freedom, complex non-linear dynamics, and self-occlusion in visual perception. The key difficulty is a suitable representation, rich enough to capture the object shape, dynamics for manipulation and yet simple enough to be estimated reliably from visual observations. This work aims to learn latent \textit{Graph dynamics for DefOrmable Object Manipulation} (\gdoom). \gdoom approximates a deformable object as a sparse set of interacting \textit{keypoints},  which are extracted automatically from images via unsupervised learning. It learns a \textit{graph neural network} that captures abstractly the geometry and  the interaction dynamics of the keypoints. To handle object self-occlusion, \gdoom uses a recurrent neural network to track the keypoints over time and condition their interactions on the history. We then train the resulting recurrent graph dynamics model through contrastive learning in a high-fidelity simulator. For manipulation planning, \gdoom reasons explicitly about the learned dynamics model through model-predictive control applied at each keypoint. Preliminary experiments of \gdoom on a set of challenging rope and cloth manipulation tasks indicate strong performance, compared with  state-of-the-art methods. Although trained in a simulator, \gdoom transfers directly to a real robot for both  rope and cloth manipulation\footnote{Demo video available online at \url{https://youtu.be/oCfbNMx2sQI}}.
\end{abstract}

\begin{IEEEkeywords}
    Deformable object manipulation, graph neural networks
\end{IEEEkeywords}

\section{Introduction}\label{sect:intro}

Robot manipulation of rigid-body objects has made significant progress in recent years,
but many manipulation tasks in daily life involve deformable objects, \eg, pulling cables, folding clothes, or bagging groceries. Manipulating deformable objects remains an open challenge because of their large degrees of freedom, complex non-linear dynamics, and self-occlusion in visual perception. 

Early work on deformable object manipulation uses predefined visual features to capture the object shape and plans robot actions with handcrafted approximate dynamics models~\cite{miller2012geometric,saha2007manipulation,torgerson1988vision}. Predefined geometric features are often not robust and introduce errors in state estimation. The approximate dynamics model then compounds the error during long-horizon planning~\cite{karkus2019dan,ma2019particle}. Recent data-driven methods shun explicit modeling and learn policies that directly map raw visual observations to robot actions~\cite{matas2018sim,seita2019deep,wu2019learning}. 
Models, however, provide strong inductive bias for learning and improve generalization. 
Some recent methods use the particle representation and successfully learn dynamics models for complex non-rigid objects, \eg, fluids~\cite{li2018learning,sanchez2020learning,chen2021ab}, but
they require a large number of particles for accurate dynamics prediction. The resulting high computational cost makes them unsuitable for real-time manipulation planning. 
\begin{figure}[t]
	\centering
	\begin{tabular}{c@{\hspace{2pt}}c@{\hspace{2pt}}c@{\hspace{2pt}}c}
      \includegraphics[width=0.2\linewidth]{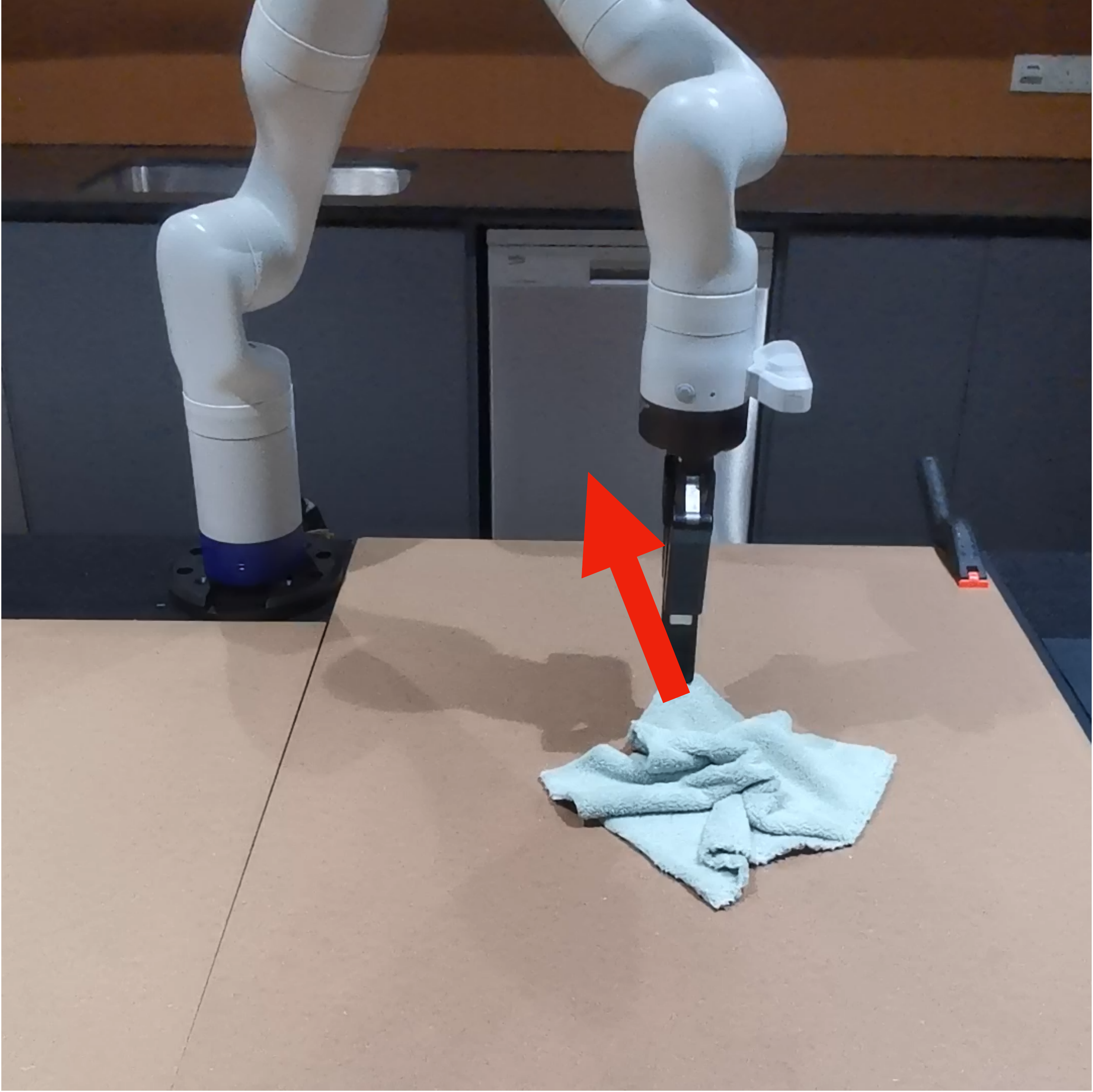} &
	\includegraphics[width=0.2\linewidth]{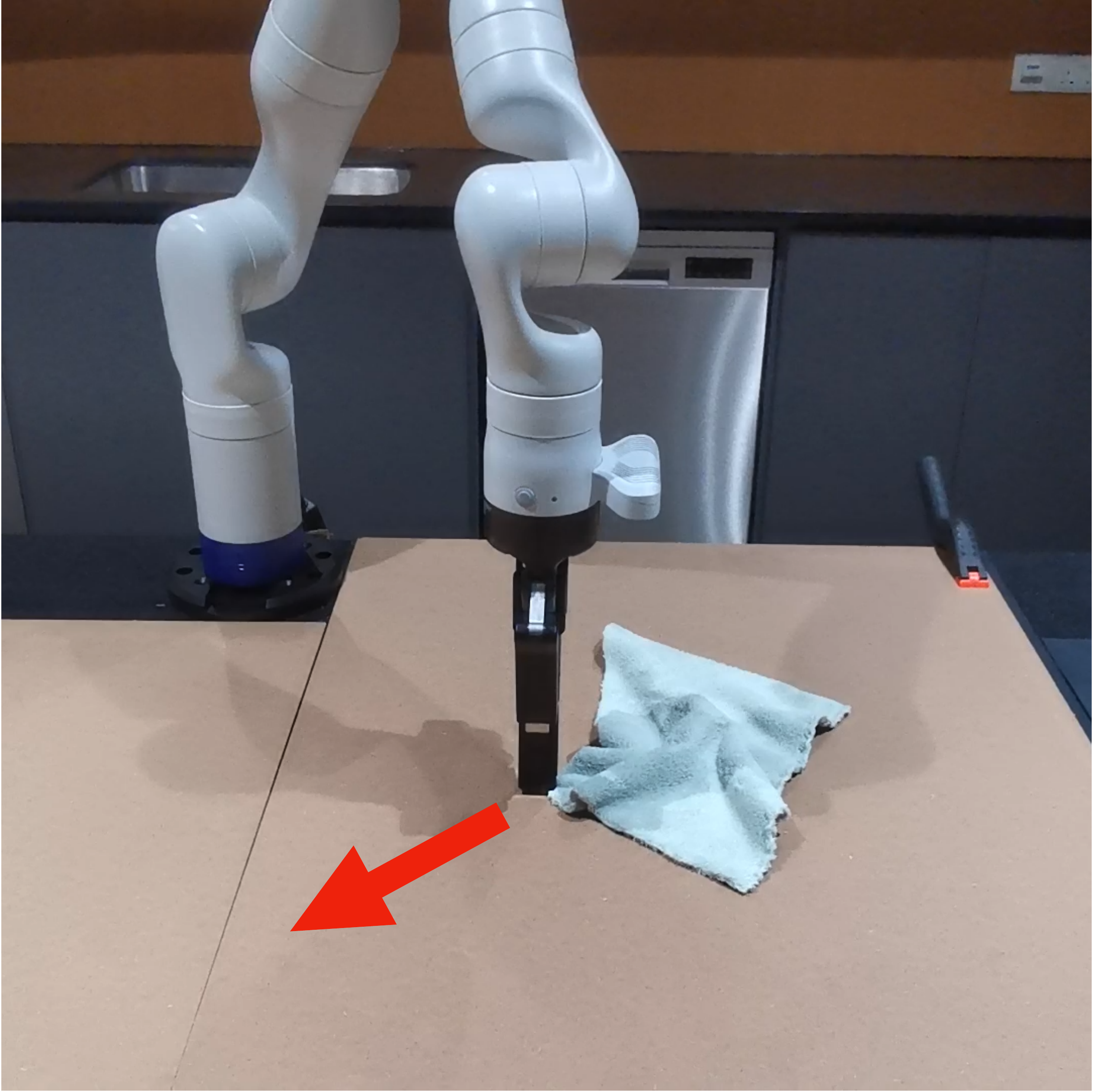} &
    \includegraphics[width=0.2\linewidth]{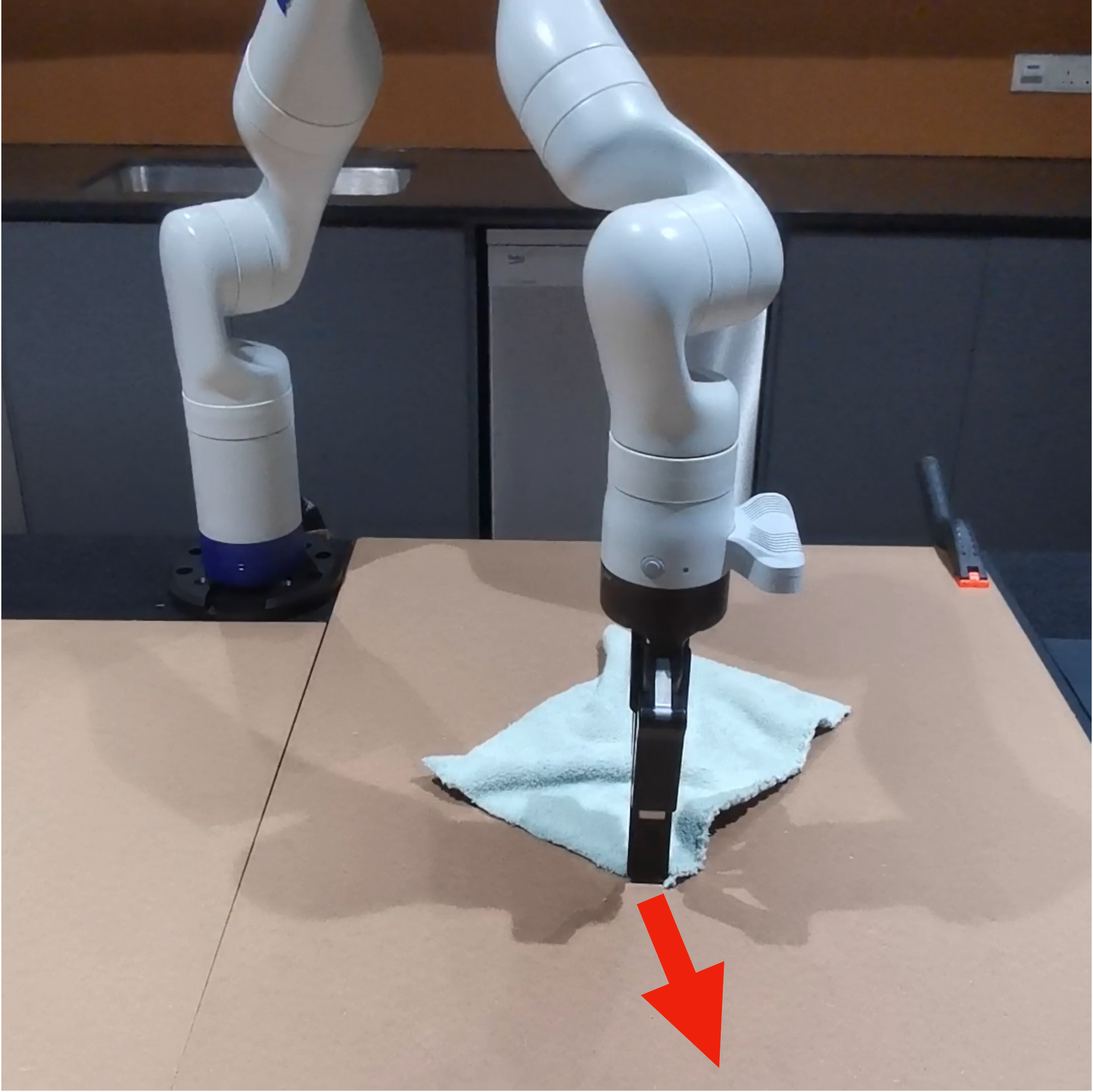} &
    \includegraphics[width=0.2\linewidth]{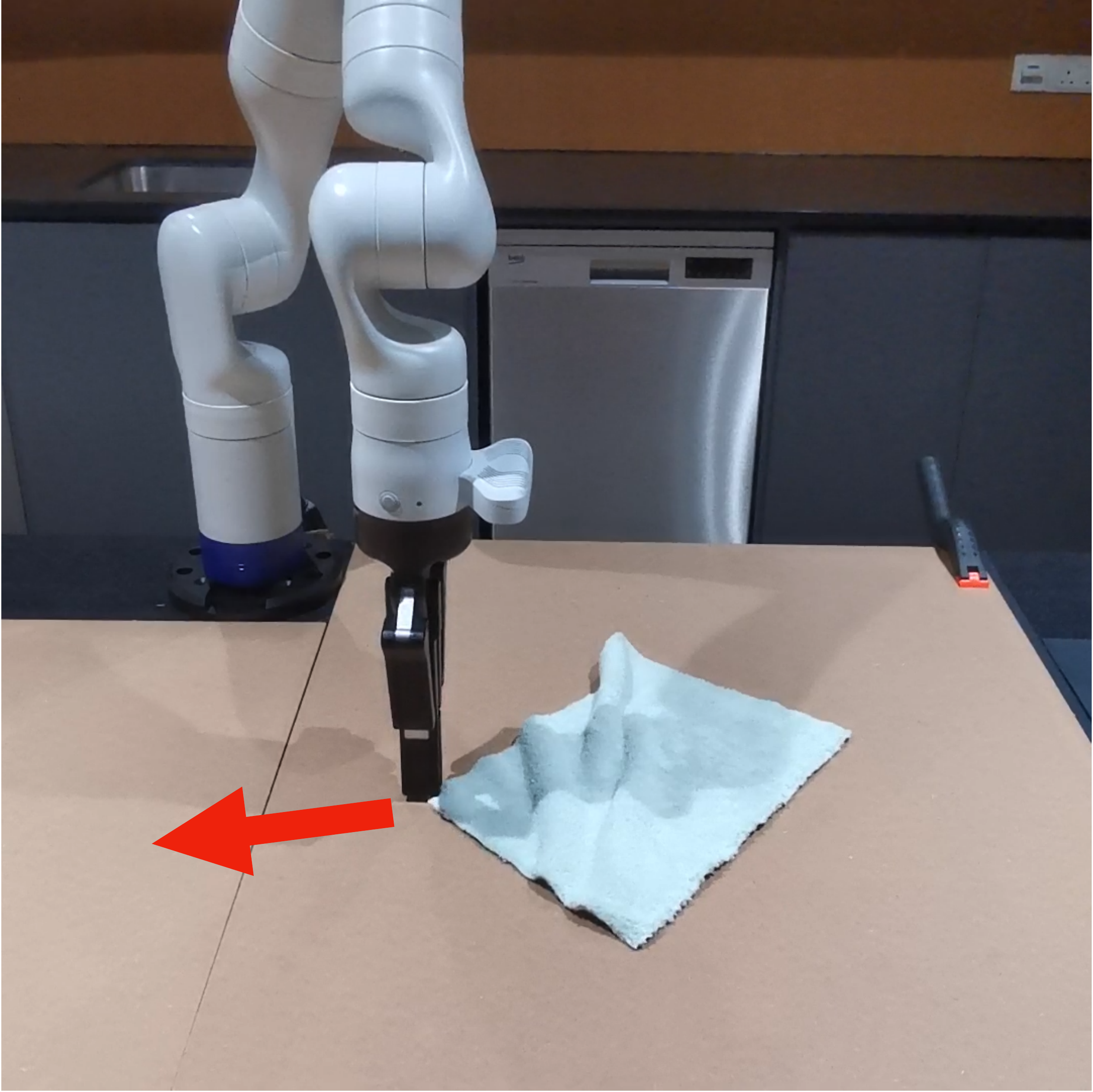} \\
   \includegraphics[width=0.2\linewidth]{figs/kp_vis/cloth0} &
	\includegraphics[width=0.2\linewidth]{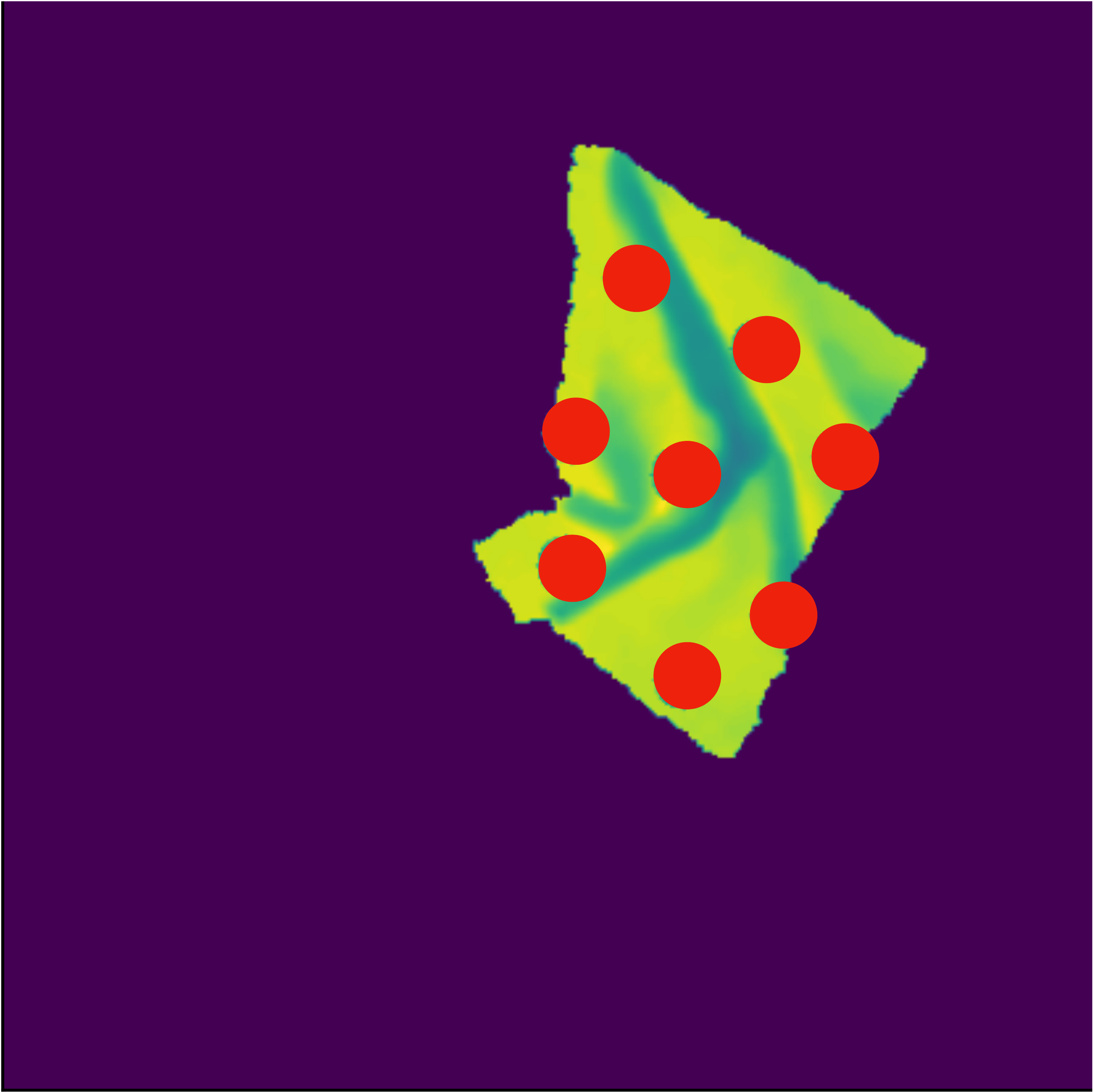} &
    \includegraphics[width=0.2\linewidth]{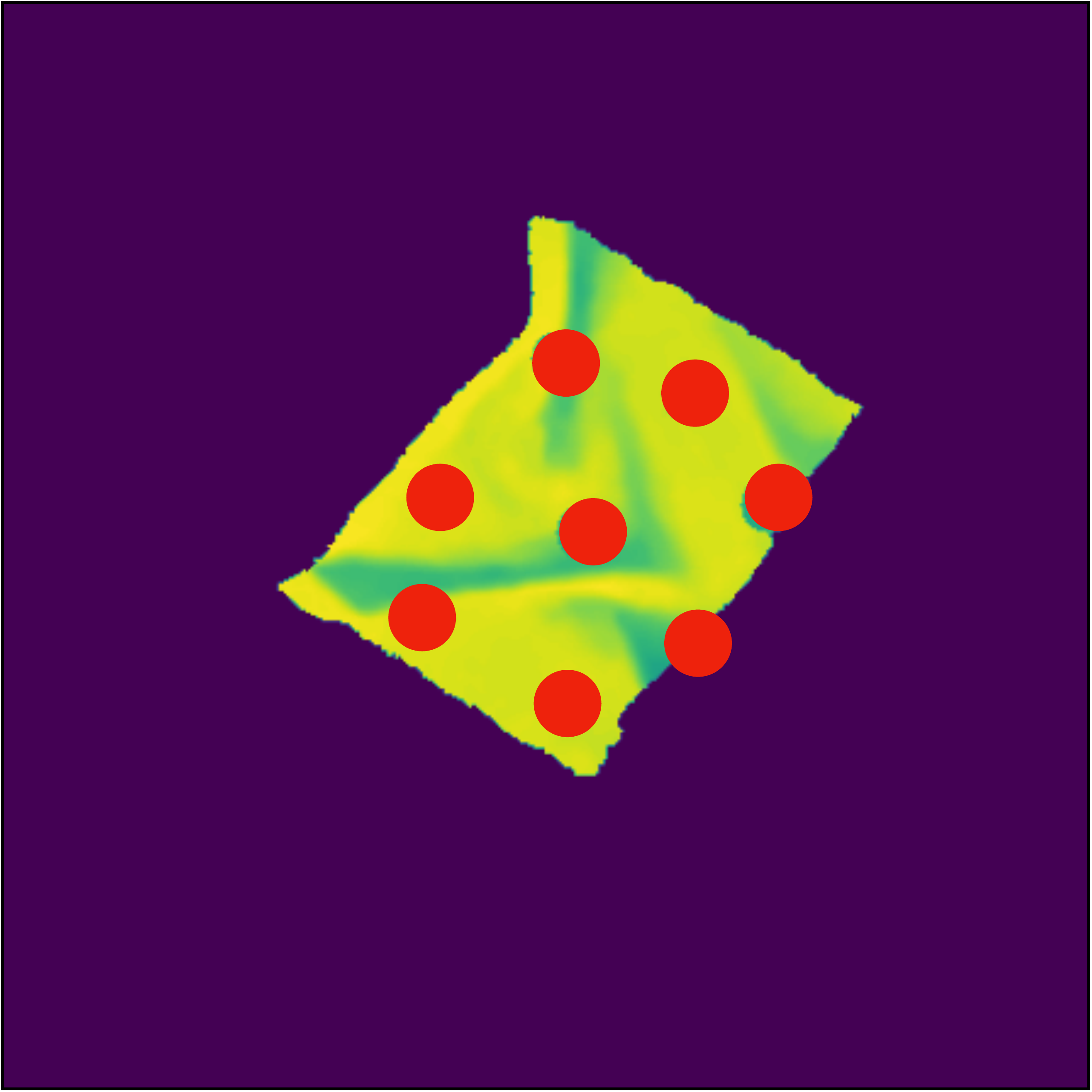} &
    \includegraphics[width=0.2\linewidth]{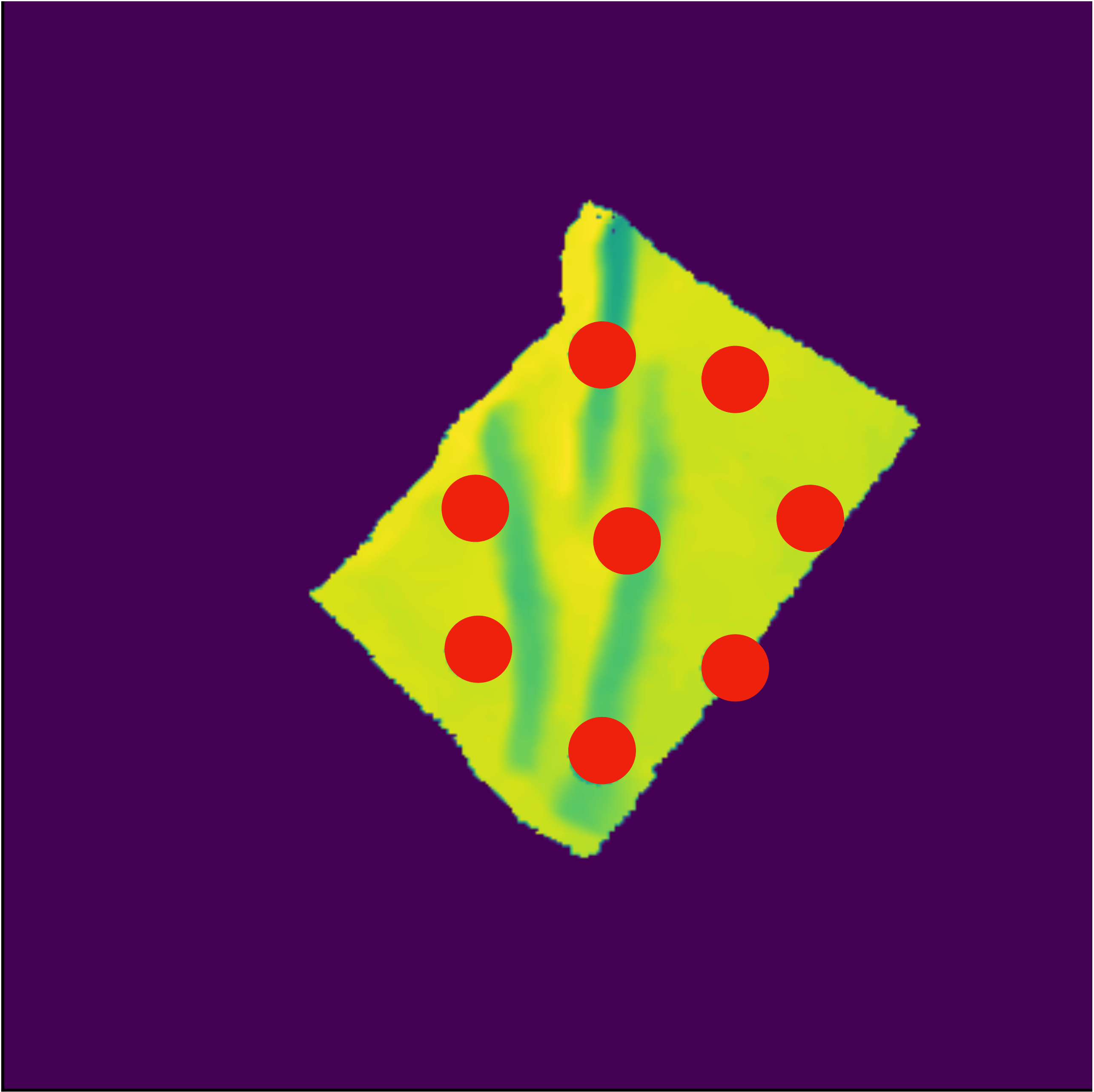} 
	\end{tabular}
	\centering
	\caption{\gdoom   flattens a piece of crumpled cloth in 4  steps, using a Kinova Gen3 robot. It reasons about a learned graph dynamics model over keypoints (bottom row), which are extracted from depth images.  The red arrows (top row) indicate robot actions. 
	}
   \vspace{-5mm}
	\label{fig:flatten_example}
\end{figure}

We hypothesize that while models of object shapes and dynamics are crucial for manipulation planning, high model accuracy may be unnecessary for many manipulation tasks in daily life. Consider how humans fold a piece of clothing. Instead of accurately reasoning the dynamics of every point on the clothing, we focus only on a few \textit{keypoints},  the collar, shoulders, \etc, which are easy to observe visually and sufficient to capture the key underlying dynamics information abstractly.  

To this end, we present  \textit{latent Graph dynamics for DefOrmable Object Manipulation} (\gdoom), a new method for visual manipulation of deformable objects. \gdoom approximates  a  deformable  object  as  a  \textit{sparse}  set  of interacting keypoints (Fig.~\ref{fig:flatten_example}). It consists of
three components. First, \mbox{\gdoom} extracts 
from depth images a set of visually-salient keypoint features,  via unsupervised learning~\cite{kulkarni2019unsupervised}.
 Second, \gdoom learns a \textit{recurrent graph neural network} model to capture the complex non-linear dynamics of keypoints. 
 It represents each keypoint as a node in a graph neural network (GNN)~\cite{wu2020comprehensive} and applies graph convolution on the keypoint features to learn ``summaries'' of the keypoint interactions abstractly.  
 To handle  object  self-occlusion,  it further makes the GNN recurrent and
conditions the keypoint interaction model on the history.  
Unlike the humans who track a set of fixed keypoints, \gdoom does not track the keypoints individually at each time step. Instead, it tracks the global statistics of the \textit{keypoint set} and uses them to reason about the dynamics of the deformable object at an abstract level. 
Finally,  G-DOOM exploits the learned  graph dynamics model and applies model-predictive control (MPC)  at each keypoint to choose the best action for manipulation.

\begin{figure*}
    \centering
    \includegraphics[width=0.9\linewidth]{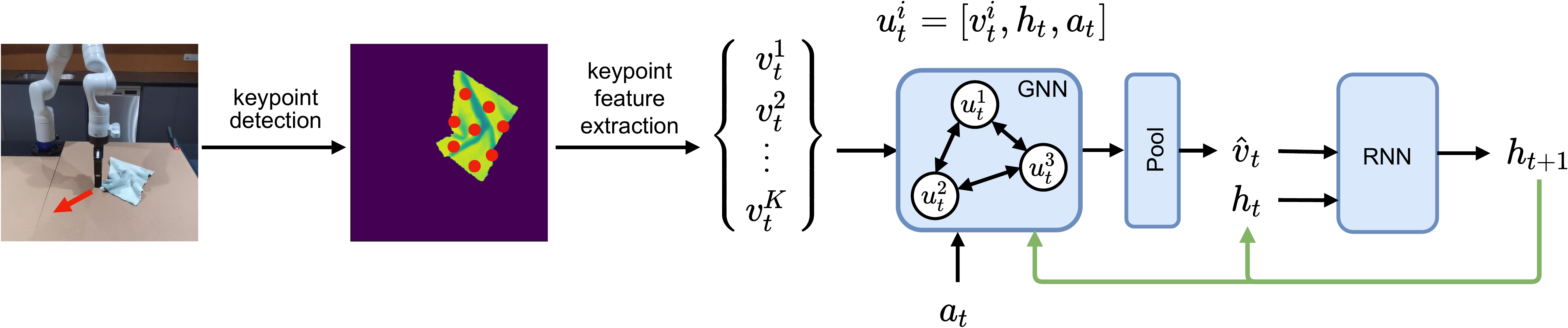}
    \caption{\gdoom performs unsupervised keypoint detection on depth images, extracts the corresponding keypoint features $\{v_t^i\}_{i=1}^K$, and compose them into a graph through learned spatial relationships. It learns recurrent graph dynamics  to predict the future states with features $\{u_t^i\}_{i=1}^K = \{[v_t^i, h_t, a_t]\}_{i=1}^K$,  which account for the spatio-temporal interactions among the local graph features $\{v_t^i\}_{i=1}^K$, an action $a_t$, and global statistics $h_t$ that captures the history. 
    }
    \label{fig:model}
\end{figure*}

We evaluated \gdoom on three deformable manipulation tasks: rope straightening, cloth flattening, and cloth folding. Experiments show that in a high-fidelity simulator, Nvidia Flex~\cite{ciccarelli2019particle}, \gdoom outperforms PlaNet~\cite{hafner2018planet} and CFM~\cite{yan2020learning}, the state-of-the-art method for visual manipulation of deformable objects.
While \gdoom is trained entirely in simulation, it transfers directly   to   a   Kinova Gen3 robot  for   both   rope   and   cloth manipulation. 
The experimental results provide the initial evidence to support our hypothesis: a sparse set of keypoints can capture the dynamics of a complex deformable object. 

\section{Related Works}

Early work on deformable object manipulation uses predefined visual features for object state estimation~\cite{li2015folding,yamakawa2011motion} and plans with handcrafted approximate dynamics models, \eg, linear models~\cite{moll2006path,saha2007manipulation,smolen2009deformation,wada2001robust}.
One major difficulty is a powerful representation that connects rich visual perception and complex object dynamics for long-horizon robot manipulation planning. 
The recent advances in data-driven methods, especially, deep learning, have brought up significant interests in learning model-free  policies~\cite{matas2018sim,mnih2016asynchronous,wu2019learning,lee2020learning,ganapathi2021learning} or  dynamics models ~\cite{hafner2018planet,ma2020contrastive,yan2020learning} from raw visual images. However, a lot of prior works on model learning use the unstructured state representation as a single vector (see, \eg, ~\cite{karkus2019dan,ma2019particle}).
Structured representations, such as particles~\cite{li2018learning,sanchez2020learning} or graphs~\cite{kipf2019contrastive,li2018learning,seita2020learning,yu2020spatio,zeng2020transporter}, generally improve the predictive power and generalization capability of the learned models. Hence, \gdoom uses a learned graph by GNNs to represent the object's dynamic state.

Our work follows the general idea of learning latent dynamics models for planning, but focuses on the structured, learnable dynamics representation over a sparse set of visual keypoints. 
Compared with model-free methods, \gdoom benefits from the structural inductive bias of the graph dynamics model for generalization.
At the same time, it is comparable in computational efficiency with methods using the simpler vector-based state representation for learning.   
Concurrent to this work, Lin~\etal~\cite{lin2021learning} propose to learn a visual connectivity graph for cloth smoothing. It represents the cloth as a dense point cloud and tries to reconstruct its 3-D geometry partially. In contrast, G-DOOM demonstrates a learned latent representation over sparse visual keypoints in the image space.
The implication is that the overall object shape rather than accurate 3-D shape reconstruction is important for manipulation, at least, in some common daily tasks.

\section{\gdoom}

\gdoom performs visual manipulation of deformable objects, using a learned graph dynamics model over a set of keypoints.  
Given a top-down image, \gdoom 
extracts visually-salient keypoint features, through
 unsupervised learning. These keypoints represent partial observations of the underlying object state. Specifically, we use depth images as the input to minimize the sim-to-real gap.  Next, to estimates the object state, \gdoom uses a learned graph dynamics model (Fig.~\ref{fig:model}) to  track a \textit{belief}, \ie, the sufficient statistics of the keypoints.
 Using the same dynamics model, \gdoom performs MPC conditioned on the detected keypoints to choose the best action.  The main components of \gdoom---learning keypoint extraction, learning the graph dynamics model, and model predictive control with the learned dynamics model--are described respectively in the three subsections below. 

\subsection{Visual Keypoint Extraction}
Since manually defining keypoints generalizes poorly to unseen configurations and objects, we leverage the Transporter Networks~\cite{kulkarni2019unsupervised} for unsupervised keypoint detection and feature extraction. The core idea of Transporter Networks is that the salient keypoints should contain sufficient information for pixel-level image reconstruction. During training, we take a pair of images sampled from the a collected dataset, $(I_{src}, I_{tgt})$, pass them through a feature extractor $f_{enc}$ and keypoint detector $f_{kp}$, which gives image features $(\phi_{src}, \phi_{tgt})$ and keypoint heatmaps $(\mathcal{H}_{src}, \mathcal{H}_{tgt})$. The method applies a \textit{transport} operation that composes the local target features $\phi_{tgt}$ around target keypoints $\mathcal{H}_{tgt}$ into the source feature map $\phi_{src}$ as $\Psi(\phi_{src}, \phi_{tgt}; \mathcal{H}_{src}, \mathcal{H}_{tgt})$. Given the reconstructed target $\hat{I}_{tgt} = f_{dec}(\Psi(\phi_{src}, \phi_{tgt}; \mathcal{H}_{src}, \mathcal{H}_{tgt}))$, we optimize the keypoint detection module by minimizing the reconstruction loss, $L_{rec} = \parallel I_{tgt} - \hat{I}_{tgt} \parallel$.

For an image $I_t$ at time $t$, we treat the keypoint heatmap $\mathcal{H}_t$ as an attention mask over the feature map $\phi_t$ and apply mean-pooling over the channel-dimension of $\phi_t$. Mathematically, the $i$-th keypoint feature is computed by $v^i_t = \textrm{MeanPool}(\mathcal{H}^i_t\phi_t)$, where $H^i_t$ denotes the $i$-th channel of the keypoint heatmap. The keypoint location $p^i_t$ can be acquired by $p^i_t = \arg\max H^i_t$. With depth images as the input, the node feature $v^i_t$ learns to extract the depth and geometric information around position $p^i_t$, which provides rich information for keypoint interaction modeling. For each image $I_t$, we construct a graph observation $G_t = (V_t, E_t)$,
where $V_t = \{v_t^i\}_{i=1}^K$ and $E_t$ is the set of edges that reflect the \textit{ground-truth} connectivity of the object. In our experiments, we show that the keypoint detector is robust and can be transferred from simulation to real-world images.

\subsection{Recurrent Graph Dynamics}
\subsubsection{Model Structure}
Given a sequence of top-down images, $I_1, I_2, \dots, I_t$, the keypoint graph prediction module outputs a sequence of graphs, $G_1, G_2, \dots, G_t$. 
Due to the sparsity of keypoints, it remains non-trivial to explicitly model the spatial keypoint interactions with a mathematical model. In addition, the self-occlusion during the deformation of the object introduces partial observability to the task, and eventually makes the exact graph matching between keypoints infeasible. 

We present \textit{Recurrent Graph Dynamics} (Fig.~\ref{fig:model}) to tackle these issues. We parameterize the high-level keypoint interactions among graph $G_t$ using \textit{Graph Neural Networks} and learn the parameters directly from data, which improves the spatial modeling performance over handcrafted models. In addition, instead of performing exact graph matching over the potentially partial keypoint graphs, we propose to extract global features from the graphs, represented by a single vector, and track the \textit{belief} $h_t$, \ie, the sufficient statistics of the global state (or we can call it the history) of the deformable object, using a recurrent neural network. Such a structure allows us to capture the informative spatial keypoint interactions and effectively estimate the global state of the object over a sequence of partial observations. At each time step $t$, we update the belief $h_t$:
\begin{align}
   {v'}_{t}^i &= \mathrm{GNN}([v_t^i, h_t, a_t], \mathrm{Nb}(v_t^i))\label{eqn:gnn}\\
   h_{t+1} &= \mathrm{RNN}(h_t, \mathrm{Pool}(\{{v'}_{t}^i\}_{i=1}^K))\label{eqn:rnn}
\end{align}
where $\mathrm{Nb}(v_t^i)$ defines the neighbor nodes of $v_t^i$ in graph $G_t$, Pool is the global pooling operation, and $a_t$ is the action taken. Specifically, in Eqn.~\ref{eqn:gnn}, by conditioning the spatial feature learning of $G_t$ on the belief $h_t$, we are able to learn temporally meaningful spatial interactions among keypoints. 

However, the visually salient keypoints reveal no underlying connectivity of the object
and explicitly defining $E_t$ remains difficult. We propose to learn \textit{soft edges} that indicates the connectivity by a probability $w_{ij}$ through end-to-end learning to maximize the predictive accuracy. We construct a fully connected graph, and rewrite Eqn.~\ref{eqn:gnn} as
\begin{equation}
    {v'}_{t}^i = \mathrm{GNN}([v_t^i, h_t, a_t], \{(v_t^j, w_{ij})\}_{j=1}^K)\label{eqn:gnn2}
\end{equation}
In our implementation, we borrow the idea from Yu \etal.~\cite{yu2020spatio} and use TGConv, a powerful attention-based graph convolution with the powerful Transformer attention mechanisms~\cite{vaswani2017attention}. In addition, since the keypoints are sparse, standard global mean-pooling or max-pooling might not sufficiently approximate the features of the keypoint set. We augment the pooling operation by \textit{Moment-Generating Function} features~\cite{ma2020contrastive}. MGF is an alternative specification of a probabilistic distribution. Mathematically, the MGF of a variable $\mathbf{X}$ is defined as $\mathbb{E}\left[e^{\theta^\intercal\mathbf{X}}\right]$, $\theta\in \mathbb{R}^n$. MGF features consider $\theta$ as a learnable vector and we compute the MGF-pooled feature $\hat{v}_t$ as
\begin{equation}
   \hat{v}_t = \textrm{MGF}(\{{v'}_t^i\}_{i=1}^K) = \left[\frac{1}{K}\sum_{i=1}^K {v'}_t^i, \sum_{i=1}^K e^{\theta^\intercal {v'}_t^i}\right]
\end{equation}
where $[\cdot, \cdot]$ denotes the concatenation of vectors.
Using MGF features, additional higher-order moment features is learned to compensate for the inaccurate Monte-Carlo approximation.

\subsubsection{Reward Function}\label{sect:reward}
To allow planning with the learned dynamics, we predict a state-dependent reward by $\hat{r}_t = f_{r}(\hat{v}_t)$ using a single fully-connected layer $f_r$ with the MGF feature $\hat{v}_t$, which encodes an ad-hoc and expressive reward signal trained by regressing manually defined rewards. 

Alternatively, \gdoom is capable of goal-oriented manipulation. Given a goal image $I_g$, \gdoom extracts the keypoint features $\{v_g^i\}_{i=1}^K$ and construct the MGF feature $\hat{v}_g = \textrm{MGF}(\{v_g^i\}_{i=1}^K)$. We can define the reward as $r_g = - \parallel\hat{v} - \hat{v}_g\parallel_2$, where $\hat{v}$ is the MGF feature of the current state. The goal-oriented reward $r_g$ is more flexible, but less expressive given complex goals, e.g., a rope with a knot. In our experiments, we mainly focus on $\hat{r}_t$.

\subsubsection{Model Learning}
G-DOOM is learned by optimizing a loss function, $L = L_{D} + \alpha L_{R}$, where $L_D$ is the dynamics loss, $L_R$ is the reward loss for $\hat{r}_t$, and $\alpha$ is a hyper-parameter balancing the two objectives.

Given the observation sequence $I_{1:t}$, we unroll the model for $T-t$ steps and predict the graph sequence $\{\hat{v}_{t+1:T}^i\}_{i=1}^K$ using Eqn.~\ref{eqn:rnn}. For the reward part, we simply train the reward predictor by minimizing the prediction error along the trajectory with $L_{R} = \sum_{t'=t+1}^T (\hat{r}_{t'} - r_{t'} )^2$.
For dynamics part, given the observation sequence $I_{1:t}$, we minimize the distance between the predicted graphs and the encoded graphs $\{v_{t+1:T}^i\}_{i=1}^K$ from the future observations $I_{t+1:T}$. To avoid exact graph matching, we instead minimize the distance between the single vector representations, \ie, the MGF-pooled encoded state $\hat{v}_t$ and MGF-pooled encoded state ${v}_t$.
Similar to \textit{Contrastive Forward Model} (CFM)~\cite{yan2020learning}, we adopt a contrastive learning objective to improve the robustness against noisy and complex observations. Different from the standard InfoNCE loss~\cite{oord2018representation}, we use an energy-based hinge-loss $L_D$ to improve the robustness of the learned representation against sparse and potentially noisy keypoint detections
\begin{equation}
   L_{D} = \sum_{t'=t+1}^T \biggl\{ \parallel \hat{v}_{t'} - v_{t'} \parallel_2 + \sum_{n=1}^N \max(0, \gamma - \parallel \hat{v}_{t'} - v_n \parallel_2) \biggr\} \label{eqn:loss}
\end{equation}
where $\{v_n\}_{n=1}^N$ are negative states encoded from a set of observations $\{I_n\}_{n=1}^N$ sampled randomly from the datasets. 

\subsection{Model-Predictive  Control with Learned Graph Dynamics}\label{sect:gcem}

Throughout our tasks, we use a pick-and-place action, $a = (x_s, y_s, x_g, y_g)$, where $(x_s, y_s)$ is the pick position and $(x_g, y_g)$ is the place position.
Inspired by the human cloth folding, we observe that an effective action normally takes effect around the keypoints. 
We introduce a simple yet effective strategy, graph-based Model-Predictive Control (MPC), where we combine the stand standard MPC with the hind-sight optimization~\cite{javdani2015shared}. The intuition behind is that we initialize $K$ action sequences $\{\mathbf{a}^i\}_{i=1}^K$, centered at the keypoint $p_t^i = (x_t^i, y_t^i)$.  In implementation, we initialize the search with $K$ hidden states $\{h_t^i\}_{i=1}^K$, unroll the trajectory with actions $\{\mathbf{a}^i\}_{i=1}^K$ and the learned dynamics (Eqn.~\ref{eqn:rnn} and Eqn.~\ref{eqn:gnn2}), and maximize the predicted reward $\{\hat{R}^i = \sum\limits_{t'=t}^{t+T}\hat{r}_{t'}\}_{i=1}^K$. The final action is acquired by $\mathbf{a}^* = \arg\max_{\mathbf{a}^{i}} \hat{R}^{i}$. This focuses the search space around the important space and empirically, we can observe a significant performance gain in our experiments.

\section{Experiments}

We first evaluate the proposed \gdoom on a set of rope straightening and cloth manipulation tasks in a high-fidelity simulator, NVidia-Flex~\cite{ciccarelli2019particle}. To minimize the sim-to-real gap, we use masked depth images as the input, and we show that our learned dynamics model transfers directly to a real robot.

We compare \gdoom with the state-of-the-art (SOTA) model-based deformable object manipulation method, \textit{Contrastive Forward Model} (CFM)~\cite{yan2020learning}, and a SOTA general-purpose model-based RL method, PlaNet~\cite{hafner2018planet}.
For all baselines, we use publicly available implementations. For a fair comparison, we adapt the original goal-oriented CFM to reward-driven by adding an additional reward predictor.

\begin{figure}[t]
	\centering
	\begin{tabular}{c@{\hspace*{20pt}}c}
      \includegraphics[width=0.44\linewidth]{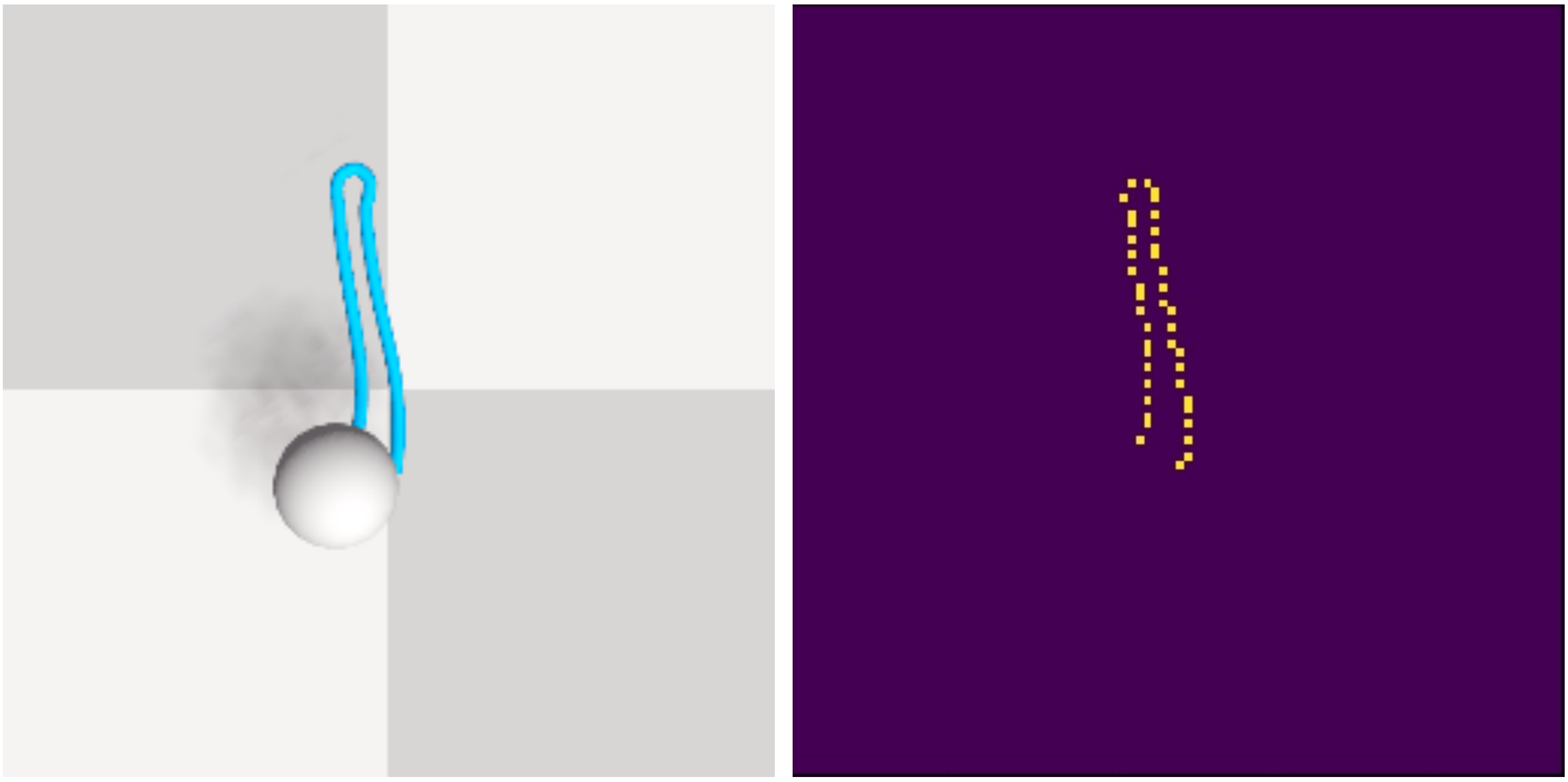} &
	\includegraphics[width=0.44\linewidth]{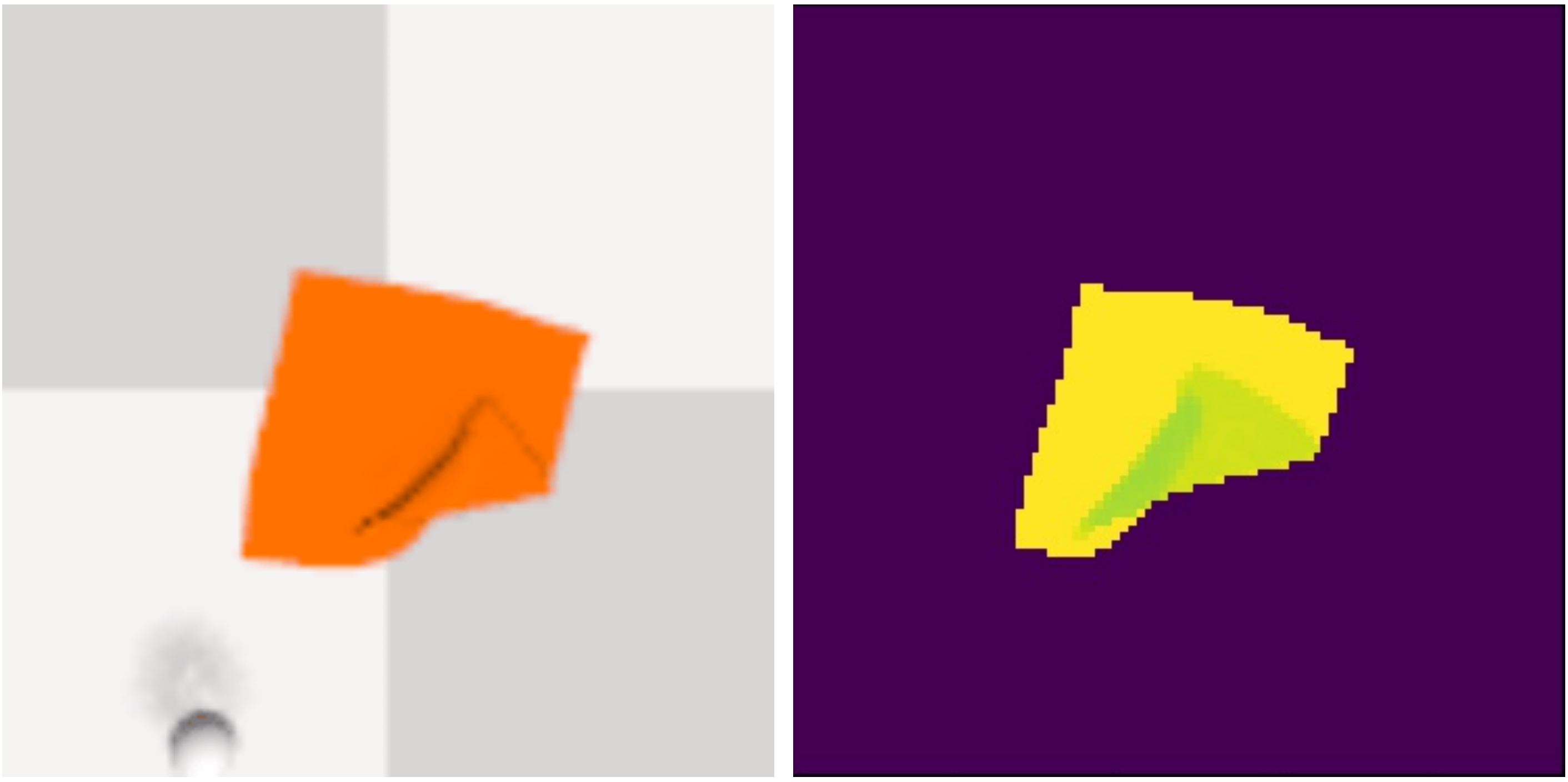} \\
	\end{tabular}
	\centering
	\caption{Deformable object simulation  in NVidia-Flex~\cite{ciccarelli2019particle}, a high-fidelity particle-based simulator. For each object,  both the 3-D visualization  and the corresponding depth image  are shown. 
	}
   \vspace{-5mm}
	\label{fig:simulator}
\end{figure}
\subsection{Experiment Setup}
\subsubsection{Simulation Environment}
We build our simulator upon SoftGym~\cite{lin2020softgym}, a soft-body simulator based on NVidia-Flex~\cite{ciccarelli2019particle}, the state-of-the-art (SOTA) particle-based simulator. Example images are shown in Fig.~\ref{fig:simulator}. We make certain modifications on SoftGym to better match the real-robot scenarios, where:
\begin{itemize}
    \item \textit{Observation space}. The observation is a depth image of size $96\times 96$ with depth values, indicating the distance to a top-down simulated camera. Specifically, we perform object segmentation and set the background values to 0. 
    \item \textit{Action space}. As mentioned in Sect.~\ref{sect:gcem}, we use a pick-and-place action, $a_t = \left[x_s, y_s, x_g, y_g\right]$, where $(x_s, y_s)$ is the pick position and $(x_g, y_g)$ is the place position. 
\end{itemize}

\begin{table*}[t]
   \parbox{.5\linewidth}{
      \caption{Simulation Experiments: Success Rate (\%)}
      \label{tab:reward_sim}
      \centering
      \begin{tabular}{lcccccc}
   \toprule
   \multicolumn{1}{l}{\multirow{2}{*}{}} & \multicolumn{4}{c}{Rope}                                              & \multicolumn{2}{c}{Cloth}        \\
   \midrule
   \multicolumn{1}{l}{}                  & 0$^\circ$               & 45$^\circ$              & 90$^\circ$              & 135$^\circ$             & Flatten        & Fold            \\
   \cmidrule(lr){1-1}\cmidrule(lr){2-5}\cmidrule(lr){6-7}
   CFM                                   & 24.7          & 18.4          & 9.1           & 11.6          & 29.2          & 59.6          \\
PlaNet                                & 39.1          & 68.1          & 64.4          & 82.4          & 35.9          & 73.2          \\
\gdoom                                & \textbf{65.7} & \textbf{80.3} & \textbf{73.2} & \textbf{98.1} & \textbf{95.1} & \textbf{92.5}\\
   \bottomrule
   \end{tabular}
   }
   \parbox{.5\linewidth}{
      \caption{Simulation Experiments: Predictive Dynamics Accuracy (\%)}
      \label{tab:dyna_acc}
      \centering
      \begin{tabular}{lcccccc}
   \toprule
   \multicolumn{1}{l}{\multirow{2}{*}{}} & \multicolumn{4}{c}{Rope}                                              & \multicolumn{2}{c}{Cloth}        \\
   \midrule
   \multicolumn{1}{l}{}                  & 0$^\circ$               & 45$^\circ$              & 90$^\circ$              & 135$^\circ$             & Flatten        & Fold            \\
   \cmidrule(lr){1-1}\cmidrule(lr){2-5}\cmidrule(lr){6-7}
   CFM    & 47.7          & 44.8          & 48.4          & 49.6          & 59.4          & 61.1          \\
   PlaNet & -             & -             & -             & -             & -             & -             \\
   \gdoom & \textbf{99.9} & \textbf{99.9} & \textbf{99.8} & \textbf{99.9} & \textbf{99.9} & \textbf{99.9} \\
   \bottomrule
   \end{tabular}
   }
\end{table*}

\subsubsection{Tasks} We evaluate the proposed \gdoom on 2 cloth manipulation and 4 rope straightening tasks. 
\begin{itemize}
    \item \textit{Rope Straightening}. The task is to straighten a rope and position it at the center of the image with different orientations ($0^{\circ}, 45^\circ, 90^\circ, 135^\circ$). The challenge is to understand the complex non-linear dynamics of a deformable object. 
    
    \item \textit{Cloth Manipulation}. The cloth manipulation tasks are more challenging than the rope straightening because of the partial observability caused by self-occlusions.
    It consists of two sub-tasks: (1) cloth flattening, where the task is to flatten a piece of randomly crumpled cloth. 
    (2) cloth folding, where the task is to fold a piece of randomly positioned flattened cloth in half. 
\end{itemize}

\subsubsection{Model Learning}
We split the training of \gdoom into two phases. Firstly, we pre-train the Transporter network. Qualitatively, we find using $K=8$ keypoints for cloth manipulation and $K=3$ keypoints for rope straightening performs well. Next, we freeze the weights of the Transporter and train the dynamics model. For a fair comparison, we train all baselines with the same number of epochs. 

\subsubsection{Evaluation Setups}
In all tasks, we allow a maximum sequence length of 20 steps and report the success rate over 1000 different random seeds. We define the success criteria as the distance to goal given the ground-truth particle locations in the simulator.
For the \textit{contrastive learning methods} (\gdoom and CFM), we also report the top 1 prediction accuracy~\cite{kipf2019contrastive}. This is computed by unrolling the dynamics model for 20 steps, and measure the prediction error $d(\hat{s}_{t}, s_t)$ between the predicted state $\hat{s}_t$ and the encoded state $s_t$, against $N$ negative samples $\{s_j\}_{j=1}^{J}$. The top 1 accuracy is computed by $Acc = 1/N \sum_{n=1}^N \mathbb{I}(\forall j, d(\hat{s}_t^n, s_t^n) < d(\hat{s}_t^n, s^j))$.

\subsection{Simulation Experiment}
We present the highest reward achieved in Tab.~\ref{tab:reward_sim} and the dynamics accuracy for in Tab.~\ref{tab:dyna_acc}. We observe that:
\begin{figure*}[t]
	\centering
	\begin{tabular}{c@{\hspace{25pt}} c}
      \includegraphics[width=0.235\linewidth]{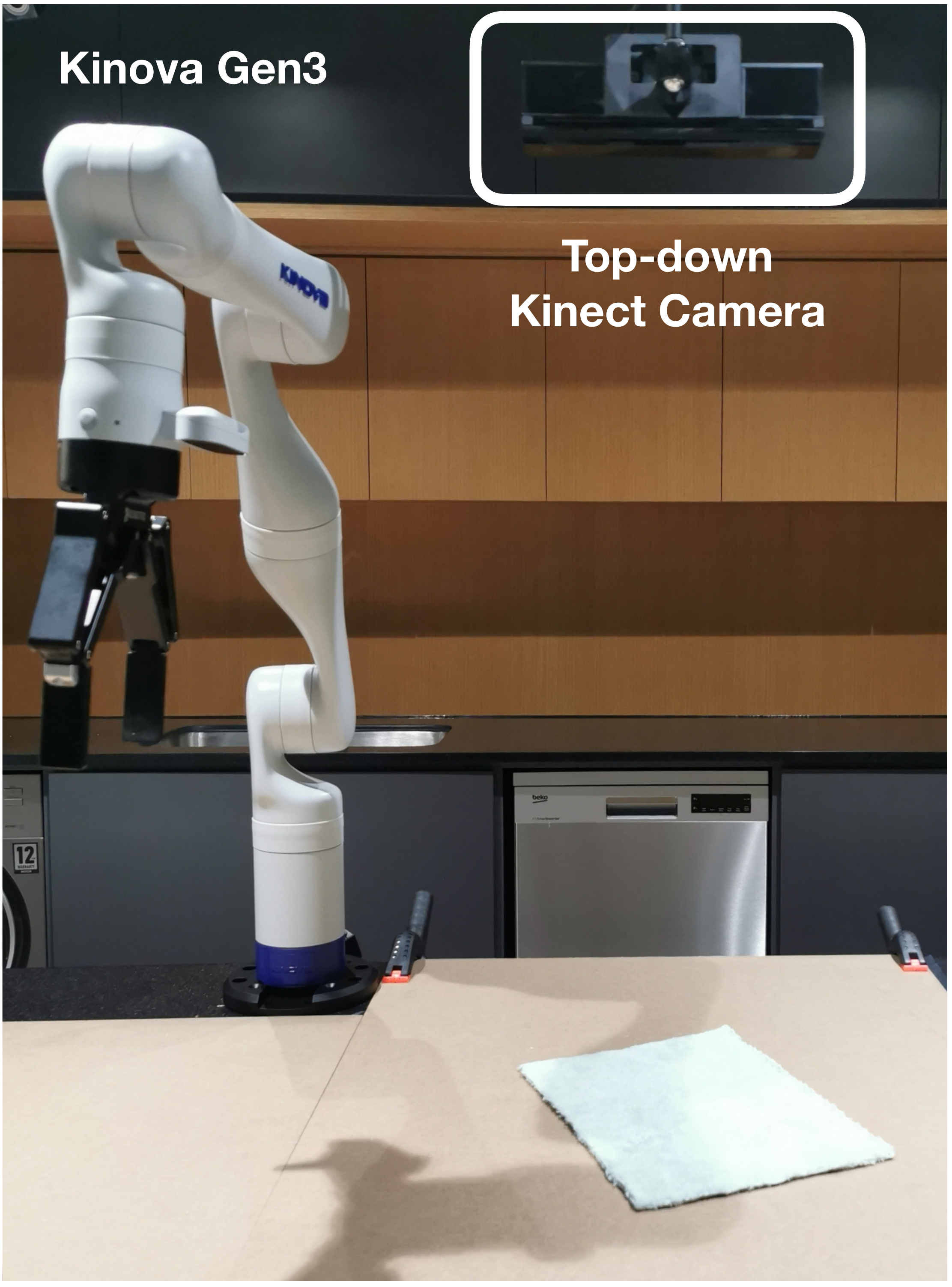} &
    \includegraphics[width=0.65\linewidth]{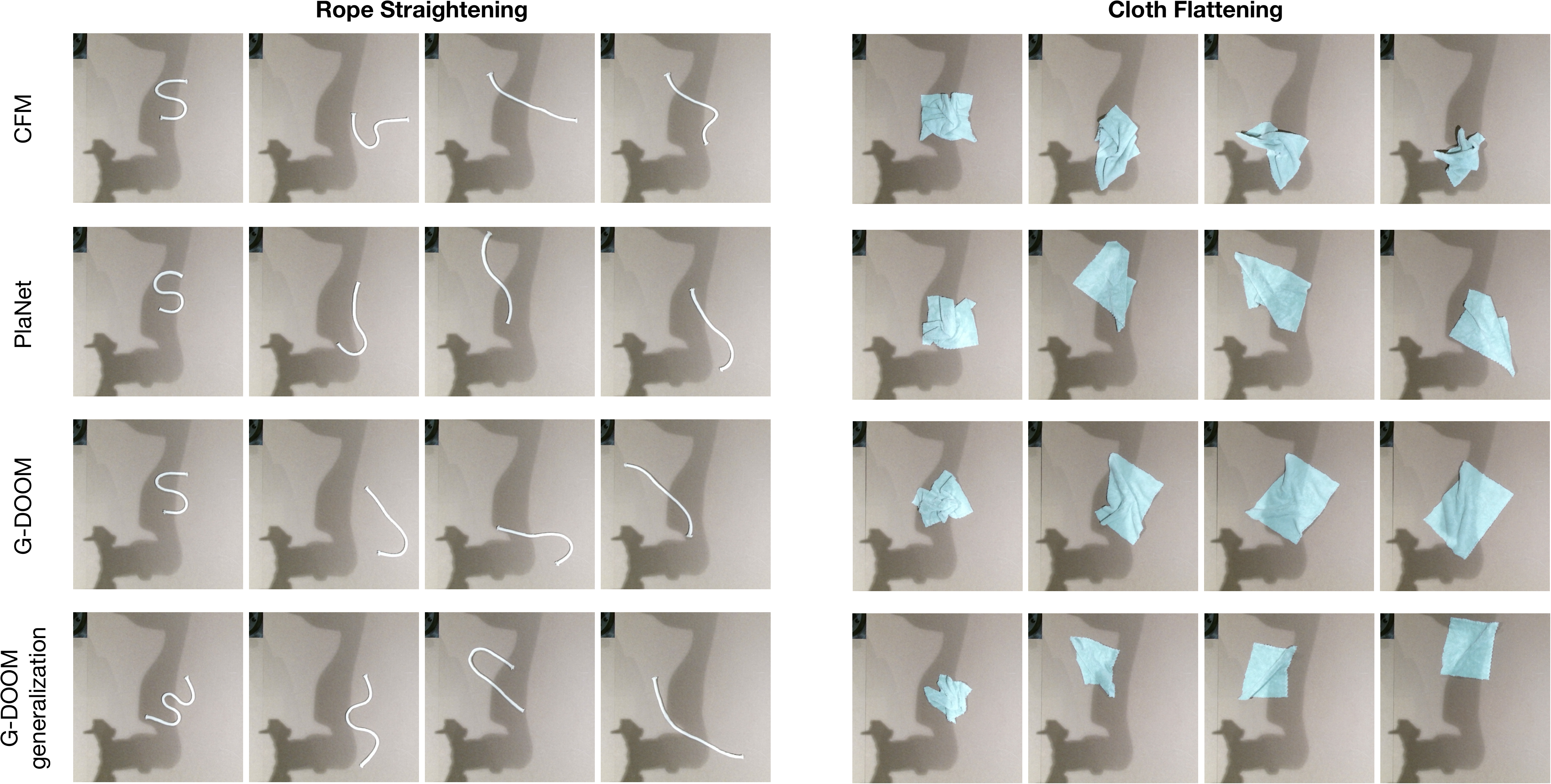} \\
    (\subfig a) & (\subfig b)
	\end{tabular}
	\centering
	\caption{Robot experiments on rope straightening ($135^\circ$) and cloth flattening. (\subfig a) We use a Kinova Gen3 robot with a Robotiq hand for manipulation and a Kinect 2.0 camera with a top-down view for depth sensing.  (\subfig b) We show the execution trajectory of each method (CRM, PlaNet, and G-DOOM) on the two tasks over 4 time steps.  
	\gdoom generalization (bottom row) shows that the learned G-DOOM model generalizes to different initial configurations or different object variants, \eg, a longer rope and a smaller piece of cloth.
	}
   \vspace{-5mm}
	\label{fig:robot_vis}
\end{figure*}

\textit{Graph dynamics better captures complex dynamics.} In all cases, \gdoom achieves the highest success rate. It suggests that the graph-based dynamics generally improve the quality of the learned dynamics model and the overall performance, compared to the single-vector-based dynamics. Besides, the advantage of \gdoom is clearer on cloth manipulation tasks than the simpler rope straightening tasks. This further emphasizes the benefit of graph structure in the dynamics.

\textit{Belief tracking is necessary for complex dynamics.} Both PlaNet and \gdoom perform belief tracking with an additional hidden vector using RNNs, while CFM is state-dependent, \ie, remembers no history. 
In most of the cases, CFM performs worse than PlaNet and \gdoom, which is contradictory to the original results of the CFM paper~\cite{yan2020learning}. The reason is that originally, CFM allows a longer execution sequence length with a smaller step size: 40 steps are allowed for rope straightening and 100 steps are allowed for cloth manipulation. However, in our case, larger step size is used with a much shorter time limit: 20 steps for all tasks, which leads to more complex dynamics. Besides, the masked depth images contain fewer signals (most of the pixels are 0), which makes it difficult for pure contrastive learning methods.

\textit{\gdoom learns a more robust contrastive dynamics model.} We measure the \emph{dynamics accuracy}, which is 1 if the positive state-observation pair produces the lowest L2 distance, compared with other 50 negative samples~\cite{yan2020learning}. In Tab.~\ref{tab:dyna_acc}, CFM achieves a low dynamics accuracy, while \gdoom achieves almost 100\% accuracy. This explains the low success rate of CFM. \gdoom focuses on keypoints and learns a better dynamics model with the recurrent graph dynamics. 

\subsection{Real Robot Experiment}
We evaluate our learned model on a Kinova Gen3 robot, as shown in Fig.~\ref{fig:robot_vis}\subfig a. To collect high-quality depth images, we mount a top-down Kinect 2.0 camera over the workspace.
We observe that high-quality depth images and the simplified pick-and-place action model help to minimize the sim-to-real gap, and our trained models transfer directly to the real robot. 

\textit{Evaluation metric:} We measure the distance-to-goal by counting the number of pixels within a goal region. Denoting the set of pixels covered by a deformable object as $S_{o}$, we define the score as follows. For rope straightening tasks, we define goal region $S_g$ to be a rectangle centered in the middle of the image rotated for different degrees ($0^\circ, 45^\circ, 90^\circ, 135^\circ$), and measure $\textrm{score} = |S_o \cap S_g|$; for cloth flattening, we simply compute the total number of pixels of the covered area by $\textrm{score}=|S_o|$; for cloth folding, we define the goal area to be half of the cloth in the initial frame and measure $\textrm{score} = -\left||S_o| - |S_g|\right|$. All results are averaged over 3 random seeds.
\begin{table}[t]
   \centering
   \caption{Real Robot Experiment Results}
   \label{tab:real_robot}
   \begin{tabular}{lcccccc}
   \toprule
   \multicolumn{1}{l}{\multirow{2}{*}{}} & \multicolumn{4}{c}{Rope}                                          & \multicolumn{2}{c}{Cloth}           \\
   \midrule
   \multicolumn{1}{l}{}                  & 0$^\circ$              & 45$^\circ$             & 90$^\circ$             & 135$^\circ$            & Flatten           & Fold            \\
       \cmidrule(lr){1-1}\cmidrule(lr){2-5}\cmidrule(lr){6-7}
   
   CFM                                   & 1.378           & 33.26         & 45.64          & 14.49          & 515.24            & -158.15         \\
   PlaNet                                & 40.93          & \textbf{68.52}           & 36.92          & 11.51 & 946.82            & -199.08         \\
   \gdoom                                & \textbf{81.91} & 67.56 & \textbf{47.92} & \textbf{53.50}          & \textbf{1,458.15} & \textbf{-42.15}\\
   \bottomrule
   \end{tabular}
   \vspace{-3mm}
   \end{table}

\textit{Results:} The quantitative results of the real robot experiments are given in Tab.~\ref{tab:real_robot} and visualizations are provided in Fig.~\ref{fig:robot_vis}. 

\textit{\gdoom generally outperforms the baselines.} In real robot experiments, \gdoom achieves higher scores than the baselines, which is consistent with our simulation results. 

\textit{Graph-based dynamics allow \gdoom to generalize better.} In the simulation, PlaNet achieves reasonable performance on rope straightening tasks, while on a real robot, it fails on rope straightening $0^\circ$ and $135^\circ$. In contrast, \gdoom generalizes in all cases. This is because by down-sampling an object into a keypoint-based graph, \gdoom constructs an information bottleneck that filters the high-frequency noise and maintains a minimum amount of information for modeling the dynamics. Also, the recurrent graph dynamics compensate for the information loss. 
As shown in Fig.~\ref{fig:robot_vis}\subfig b \gdoom (generalization), our trained model can be directly applied to different objects, \eg, longer ropes and smaller cloths. 
Due to the space limit, we visualize two real-robot tasks in Fig.~\ref{fig:robot_vis}\subfig b. 

\subsection{Additional Qualitative Results}

\begin{figure}[t]
	\centering
	\begin{tabular}{c c@{\hspace{2pt}}c@{\hspace{2pt}}c@{\hspace{2pt}}c}
   \includegraphics[width=0.18\linewidth]{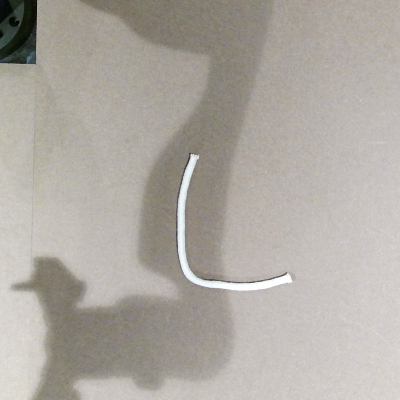} &
   \includegraphics[width=0.18\linewidth]{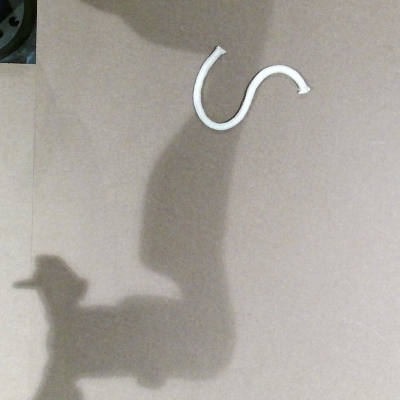} &
   \includegraphics[width=0.18\linewidth]{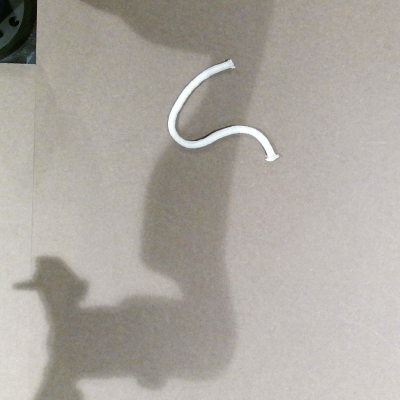} &
   \includegraphics[width=0.18\linewidth]{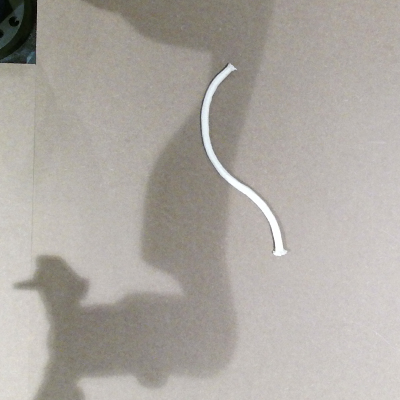} &
   \includegraphics[width=0.18\linewidth]{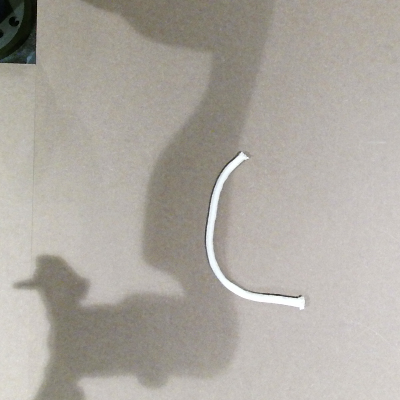} \\
   \includegraphics[width=0.18\linewidth]{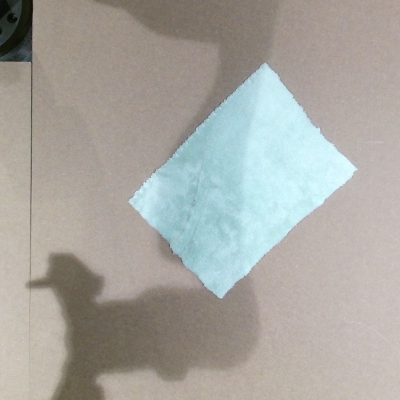} &
   \includegraphics[width=0.18\linewidth]{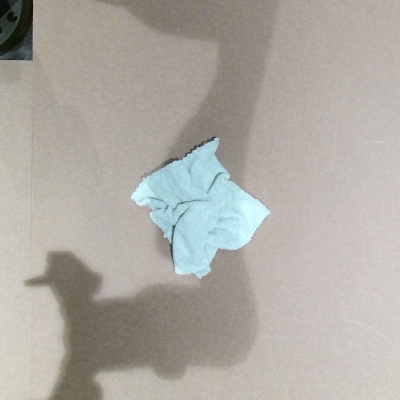} &
   \includegraphics[width=0.18\linewidth]{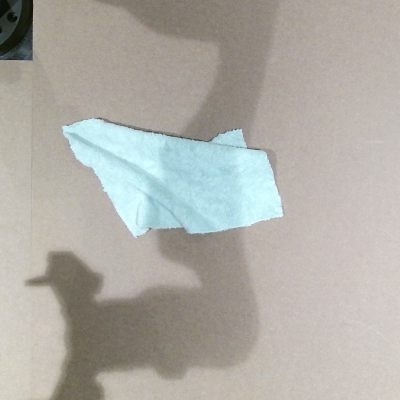} &
   \includegraphics[width=0.18\linewidth]{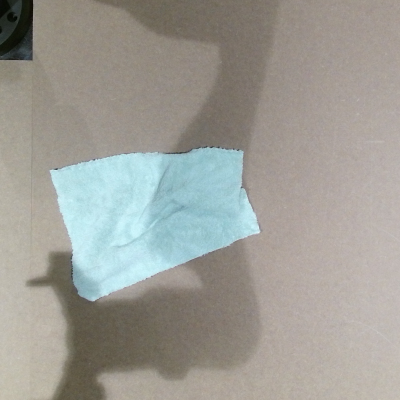} &
   \includegraphics[width=0.18\linewidth]{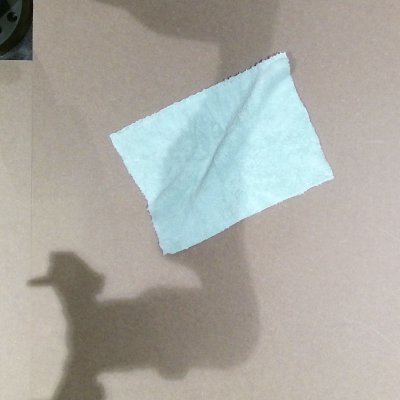} 
	\end{tabular}
	\centering
	\caption{\gdoom performs goal-directed manipulation with a goal image as the input (first column).
	}
   \vspace{-5mm}
	\label{fig:goal_oriented}
\end{figure}
\begin{figure}[t]
	\centering
	\begin{tabular}{c@{\hspace{2pt}}c c@{\hspace{2pt}}c}
      \includegraphics[width=0.18\linewidth]{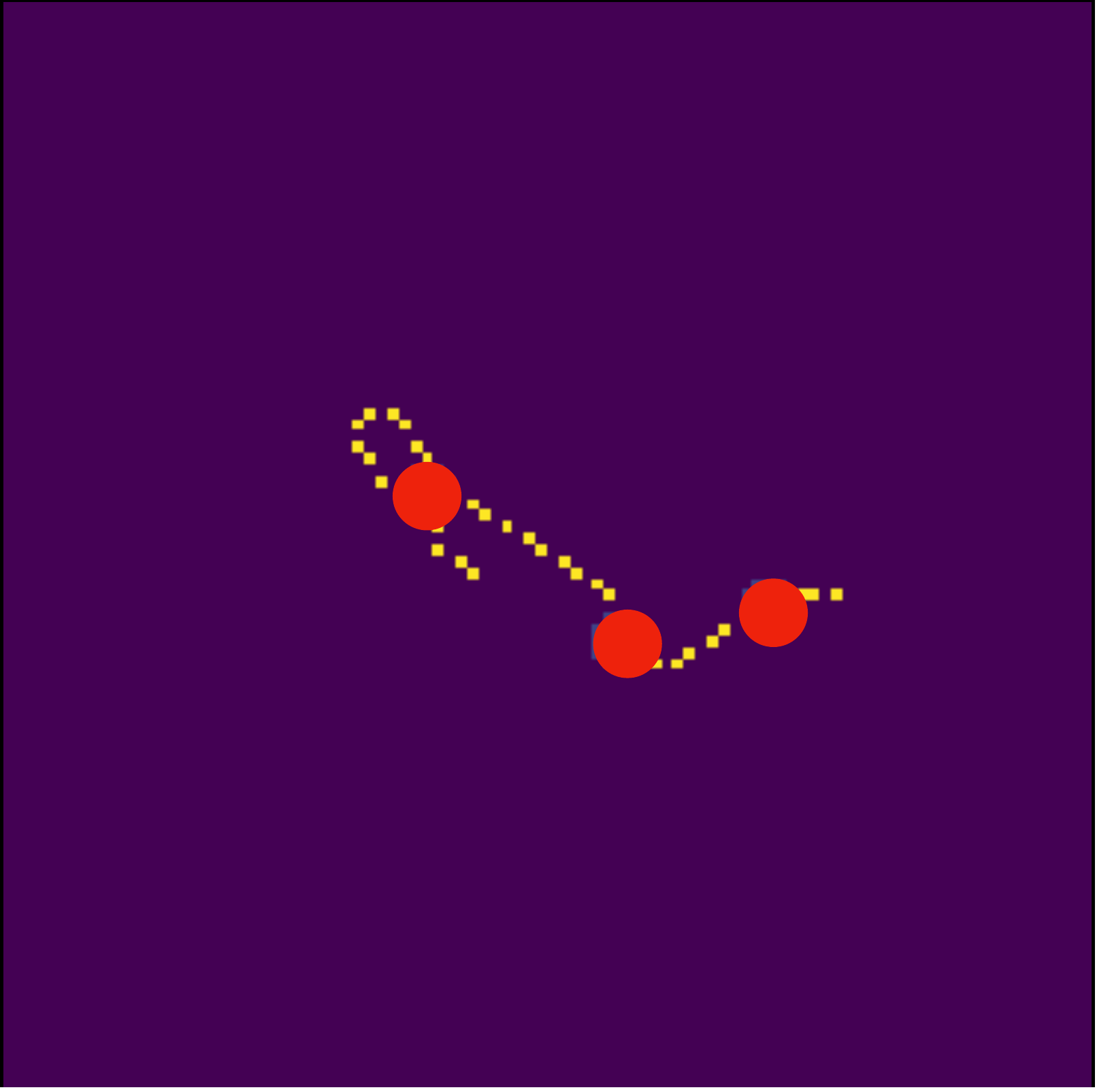} &
    \includegraphics[width=0.18\linewidth]{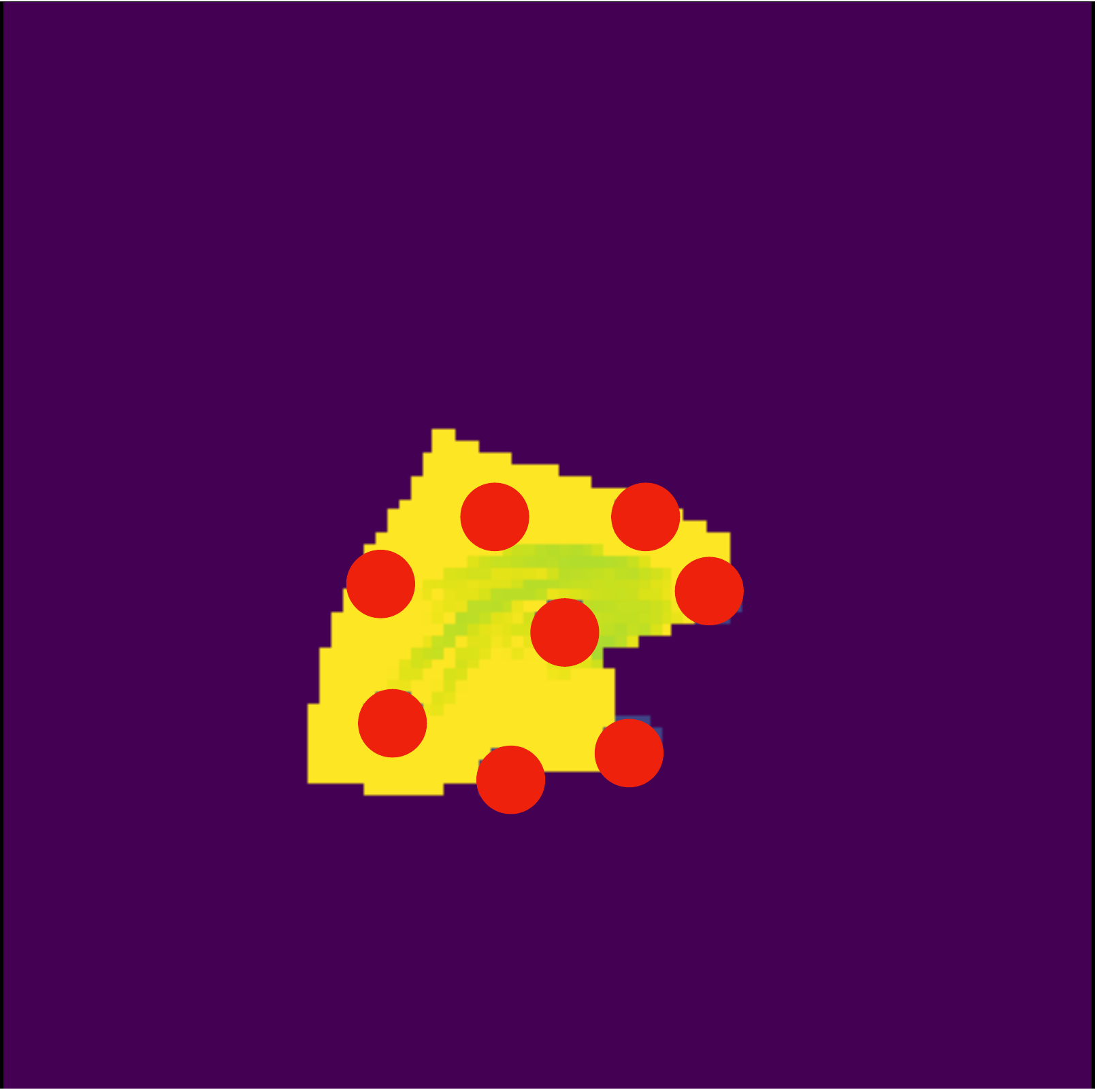} &
    \includegraphics[width=0.18\linewidth]{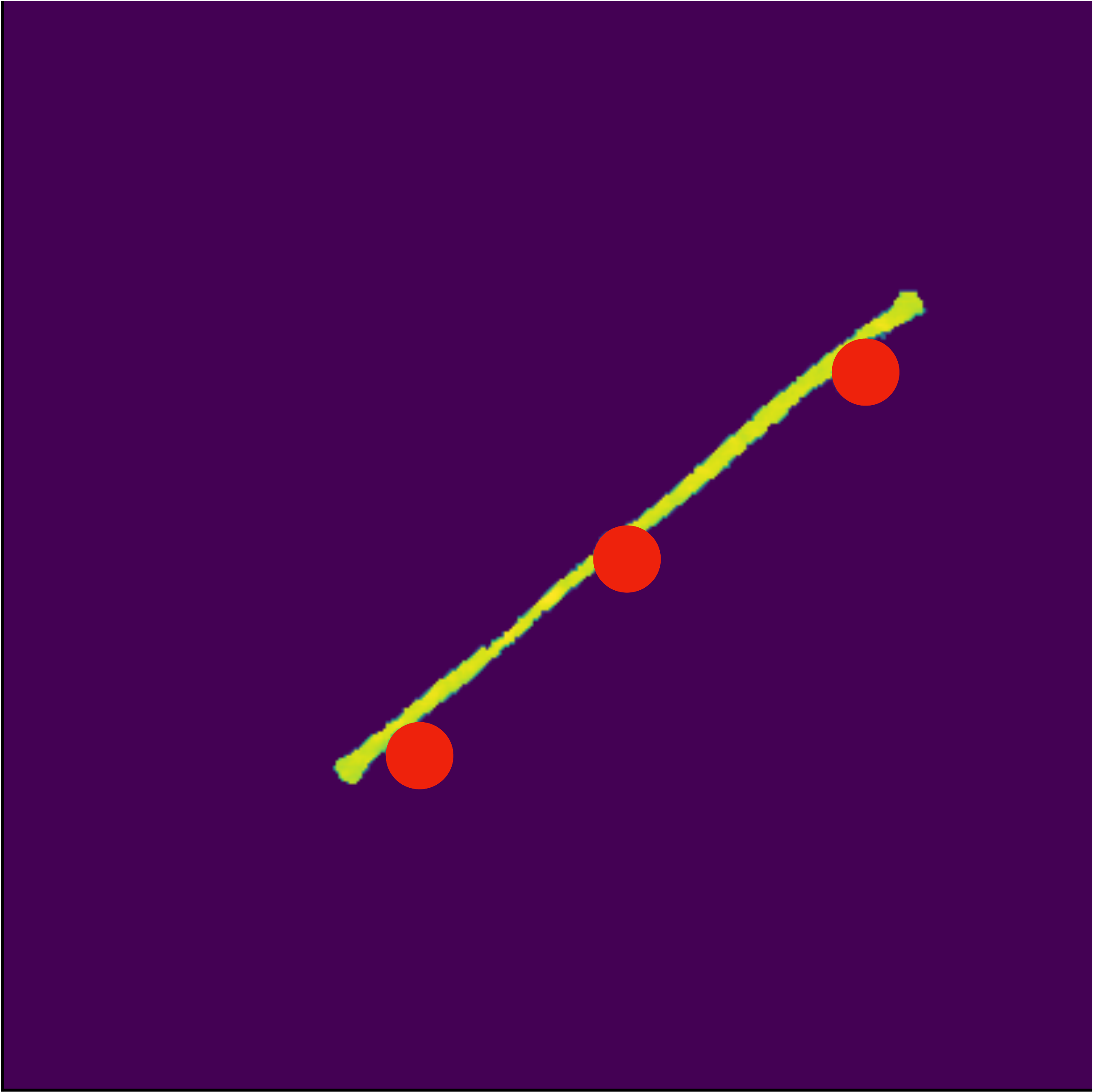} &
    \includegraphics[width=0.18\linewidth]{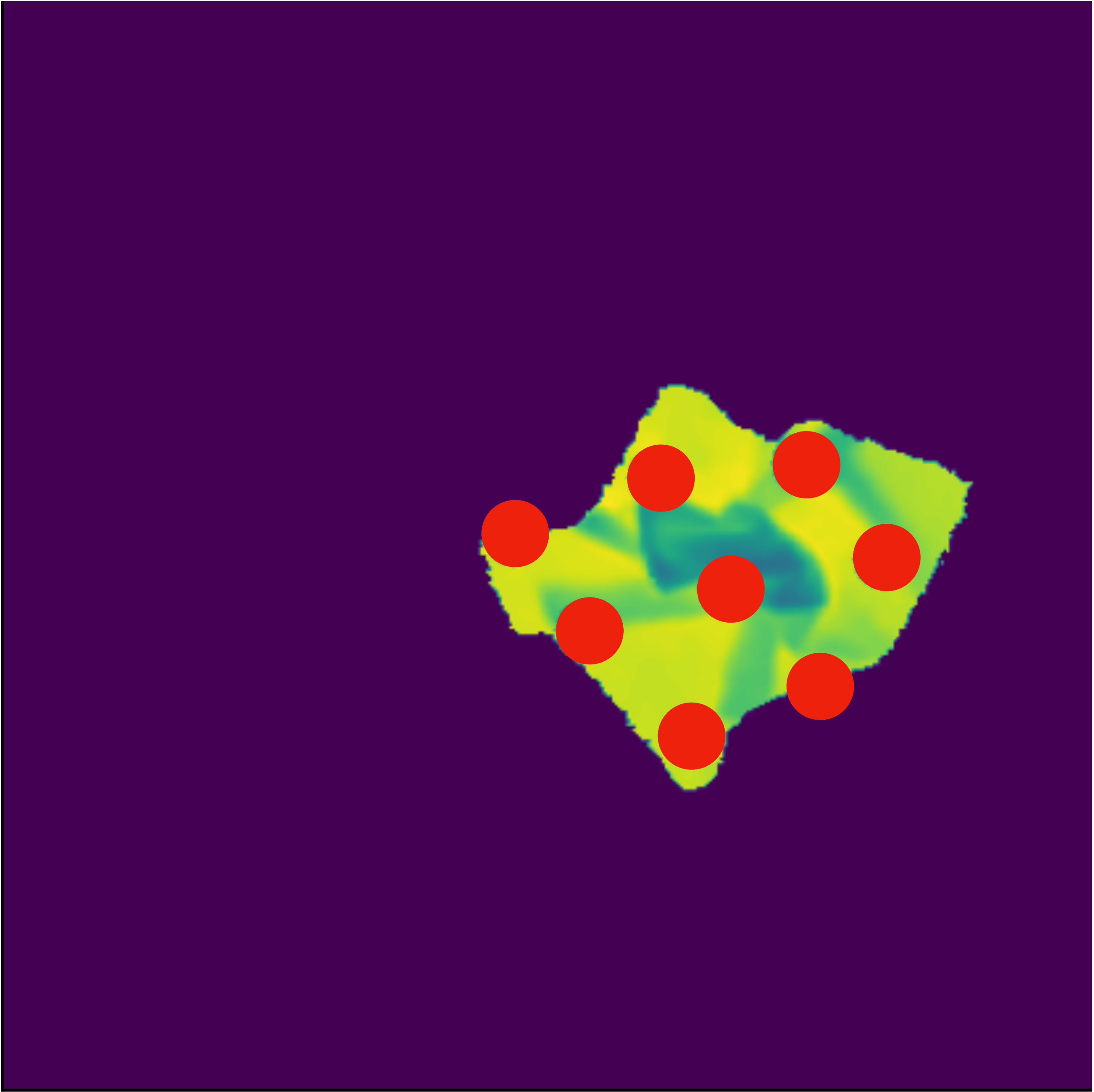} \\
   \multicolumn{2}{c}{\footnotesize simulation} & \multicolumn{2}{c}{\footnotesize real world} 
	\end{tabular}
	\centering
	\caption{Keypoint detections in simulation and real-robot experiments. 
}
	\label{fig:kpvis}
\end{figure}

As discussed in Sect.~\ref{sect:reward}, \gdoom can also be used to perform the goal-directed manipulation with the contrastively learned dynamics. We provide qualitative results of this in Fig.~\ref{fig:goal_oriented}. With an image of a ``L''-shaped rope or an image of a piece of flattened cloth as the input, \gdoom can successfully position the rope and cloth to the target configuration. However, we also noticed that it remains difficult for the goal-oriented policy to tackle tasks that require careful 3D understanding, \eg, cloth folding. This is because the 2D goal image provides no 3D understanding of the task. We leave it for future study.
We also visualize the learned keypoints (Fig~\ref{fig:kpvis}) in both simulation and  real-robot experiments. Because of the use of depth sensing, the sim-to-real gap is moderate, and we can reliably detect keypoints in both cases.  

\subsection{Ablation Study}
We conduct an extensive ablation study to understand the influence of each proposed \gdoom component (Tab.~\ref{tab:ablation}). 

\textit{Imposing a graph structure in the dynamics model generally improves the performance of the algorithm.} Compared with the standard \gdoom, the NoGraph variant removes the graph-based representation and dynamics. We observe that \gdoom generally outperforms the NoGraph variant in all tasks, which suggests that the graph-based representation better captures the dynamics of a deformable object.

\textit{Belief tracking with RNN is necessary to handle partial observability.} The NoRNN variant removes the RNN and directly uses a single TGConv as the transition function $f_{gnn}$. We observe that NoRNN achieves reasonable performance on rope straightening tasks, while on cloth manipulation where partial observability exists, its performance degrades. It suggests that belief-tracking with a RNN is useful to tackle the self-occlusions in cloth manipulation tasks.

\textit{Hinge-loss-based contrastive learning improves the robustness of the learned dynamics model.} The NoContrastive variant of \gdoom removes the contrastive component in loss function Eqn.~\ref{eqn:loss} and minimizes only the difference between the true states and predicted states. Also, the InfoNCE variant replaces the hinge loss with the standard InfoNCE~\cite{oord2018representation} used in CFM. They perform generally worse than \gdoom, which suggests contrastive learning is necessary, and hinge loss improves the robustness of the model.

\textit{Graph-based CEM can improve the search performance.} The CEM variant replaces the Graph-based CEM with a standard CEM, without using the keypoint positions as a prior to initialize the search. As a result, given the same number of optimization iterations, the standard CEM shows worse performance than the original \gdoom. 

\textit{TGConv improves the spatial modeling ability for a deformable object.} The GAT variant of \gdoom replaces the TGConv graph convolution by the widely adopted Graph Attention Networks (GAT)~\cite{velivckovic2017graph}. After replacing the TGConv with GAT, we observe that the overall performance of \gdoom generally degrades, which is because using the relatively simple GAT is insufficient to model the complex spatio-temporal microscopic interaction of a deformable object. 

\textit{MGF features improve the global feature extraction of a graph.} The MaxPool variant of \gdoom replaces the MGF features for global feature extraction by the MaxPooling technique generally used in point cloud learning literature~\cite{qi2017pointnet,qi2017pointnet++}. \gdoom with MGF features generally gives a higher success rate than the MaxPool variant.

\begin{table}[t]
\centering
\caption{Ablation Study: Success Rate (\%)}
\label{tab:ablation}
\begin{tabular}{lcccccc}
\toprule
\multicolumn{1}{l}{\multirow{2}{*}{}} & \multicolumn{4}{c}{Rope}                                          & \multicolumn{2}{c}{Cloth}           \\
\midrule
\multicolumn{1}{l}{}                  & 0$^\circ$              & 45$^\circ$             & 90$^\circ$             & 135$^\circ$            & Flatten           & Fold            \\
    \cmidrule(lr){1-1}\cmidrule(lr){2-5}\cmidrule(lr){6-7}

    NoGraph                               & 7.2           & 5.7           & 4.1           & 34.4          & 75.8          & 64.1          \\
    NoRNN                                 & 34.1          & 37.7          & 69.3          & 84.1          & 78.7          & 22.9          \\
    NoContrastive                         & 0.2           & 0.7           & 3.8           & 5.5           & 29.7          & 29.7          \\
    InfoNCE                               & 3.5           & 69.4          & 26.8          & 9.2           & 31.0          & 61.9          \\
    CEM                                   & 25.0          & 72.0          & 39.0          & 33.6          & 91.9          & 58.5          \\
    GAT                                   & 13.2          & 9.4           & 28.3          & 31.7          & 37.7          & 28.3          \\
    MaxPool                               & 28.1          & 31.9          & 29.9          & 65.5          & 94.8          & 55.7          \\
    \gdoom                                & \textbf{65.7} & \textbf{80.3} & \textbf{73.2} & \textbf{98.1} & \textbf{95.1} & \textbf{92.5}\\
\bottomrule
\end{tabular}
\vspace{-3mm}
\end{table}

\section{Conclusion}
 \gdoom performs visual manipulation of deformable objects. 
 Instead of modeling the full dynamics, it 
  learns a recurrent graph dynamics model over the keypoints, detected automatically from images, and uses the learned model to plan manipulation actions. The encouraging experimental results support an interesting finding: a sparse set of keypoints can captures the dynamics of a complex deformable object for manipulation planning.

At the same time,  \gdoom is the first attempt and currently has various limitations. The rope and cloth manipulation tasks attempted so far involve relatively simple and smooth dynamics. To tackle complex tasks, such as knot tying or Japanese T-shirt folding, we need more powerful dynamics models learned on much larger datasets as well as long-horizon planning methods.

\end{document}